\documentclass{article}

\usepackage{PRIMEarxiv}

\usepackage[utf8]{inputenc} 
\usepackage[T1]{fontenc}    
\usepackage{hyperref}       
\usepackage{url}            
\usepackage{booktabs}       
\usepackage{amsfonts}       
\usepackage{nicefrac}       
\usepackage{microtype}      
\usepackage{lipsum}
\usepackage{fancyhdr}       
\usepackage{graphicx}       

\newcommand{\model}{\mbox{\textsc{Streamline}}}

\newcommand{\figref}[1]{Fig.~\ref{#1}}
\newcommand{\tabref}[1]{Tab.~\ref{#1}}
\newcommand{\secref}[1]{Sec.~\ref{#1}}
\newcommand{\AlgRef}[1]{Algorithm~\ref{#1}}

\newcommand{\Appref}[1]{Appendix.~\ref{#1}}

\usepackage{amsmath}
\usepackage{amssymb}
\usepackage{mathtools}
\usepackage{minitoc}
\usepackage{amsthm}
\usepackage{subdef}
\usepackage{algorithm}
\usepackage{algorithmic}
\usepackage{xcolor}
\usepackage{wrapfig}

\title{STREAMLINE: Streaming Active Learning for Realistic Multi-Distributional Settings}

\author{Nathan Beck\thanks{Equal contribution}\\
University of Texas, Dallas\\
{\tt\small nathan.beck@utdallas.edu}
\And
Suraj Kothawade\footnotemark[1]\\
University of Texas, Dallas\\
{\tt\small suraj.kothawade@utdallas.edu}
\And
Pradeep Shenoy\\
Google Research India\\
{\tt\small shenoypradeep@google.com}
\And
Rishabh Iyer\\
University of Texas, Dallas\\
{\tt\small rishabh.iyer@utdallas.edu}
}

\begin{document}
\doparttoc
\faketableofcontents
\maketitle

\begin{abstract}
    Deep neural networks have consistently shown great performance in several real-world use cases like autonomous vehicles, satellite imaging, \etc, effectively leveraging large corpora of labeled training data. However, learning unbiased models depends on building a dataset that is representative of a diverse range of realistic scenarios for a given task. This is challenging in many settings where data comes from high-volume streams, with each scenario occurring in random interleaved episodes at varying frequencies. We study realistic streaming settings where data instances arrive in and are sampled from an episodic multi-distributional data stream. Using submodular information measures, we propose \model, a novel streaming active learning framework that mitigates scenario-driven slice imbalance in the working labeled data via a three-step procedure of \textit{slice identification, slice-aware budgeting}, and \textit{data selection}. We extensively evaluate \model\ on real-world streaming scenarios for image classification and object detection tasks. We observe that \model\ improves the performance on infrequent yet critical slices of the data over current baselines by up to $5$\% in terms of accuracy on our image classification tasks and by up to $8$\% in terms of mAP on our object detection tasks.
\end{abstract}

\section{Introduction}
\label{sec:Intro}
The ubiquity of deep models in a range of application areas is powered largely by the use of large labeled datasets, which have progressively increased in size in recent years. Unfortunately, procuring a large number of labeled instances can be expensive or even infeasible in complex or specialized applications. This makes the case for \textit{active learning} (AL), which aims to parsimoniously select beneficial instances for labeling from an unlabeled dataset. AL methods typically iterate over instance-selection and model-retraining steps, converging on high-quality models and compact datasets.

Many learning settings pose challenges for active learning. These include streaming settings, where (unlabeled) data arrives at a potentially high volume, forcing labeling to occur in periodic steps. Given that it is often infeasible to permanently store and iterate over the high volume of data arriving in the stream, performing active learning in such a setting becomes particularly challenging. We also identify another major challenge that is often encountered in real-life scenarios: the arriving data may be \textit{multi-distributional} in nature, and adequately representing each of these distributions -- which we also refer to as \textbf{slices} -- may be difficult within the available labeling budget. Accordingly, instrumenting existing AL methods to work in this new setting may not yield as much benefit as what could be possible.

\begin{figure}[t]
    \centering
    \includegraphics[width = \textwidth]{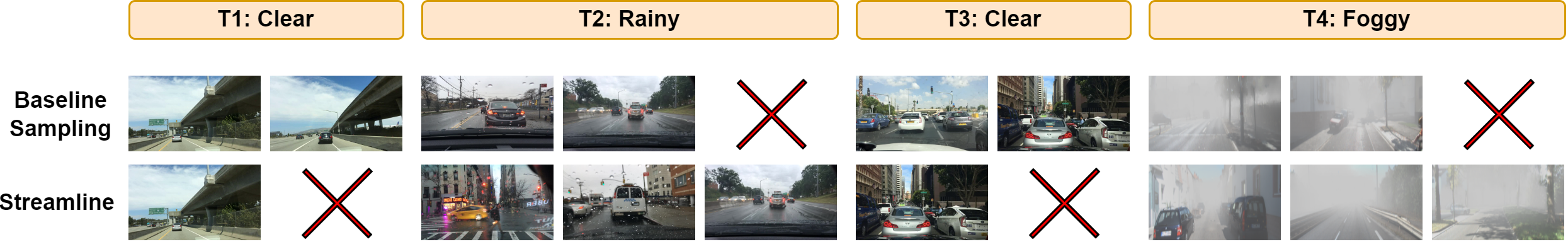}
    \caption{A real-world streaming scenario. Clear weather slices occur more frequently in comparison to the rarer rainy and foggy slices. While other basic methods sample from each slice myopically, \model\ selects fewer data points from clear slices to accumulate more budget for procuring rainy and foggy slices on which the model is under-performing.}
    \label{fig:Intro}
\end{figure}

To illustrate a real-world scenario that captures these complications, consider the setting of object detection for autonomous vehicles. Concretely, the datasets used to train such detectors are curated from data streams that originate from a fleet of cars. However, each car \emph{episodically} encounters separate driving scenarios, such as different weather conditions, different times of the day, and different regions. Each scenario corresponds to a slice of the data, and such slices arrive in episodes as dictated by each car's traversal through its environment. To complicate matters, scenarios may not even occur with the same frequency; indeed, rainy and foggy weather scenarios 
\begin{wrapfigure}{r}{0.4\textwidth}
    \includegraphics[width = 0.95\linewidth]{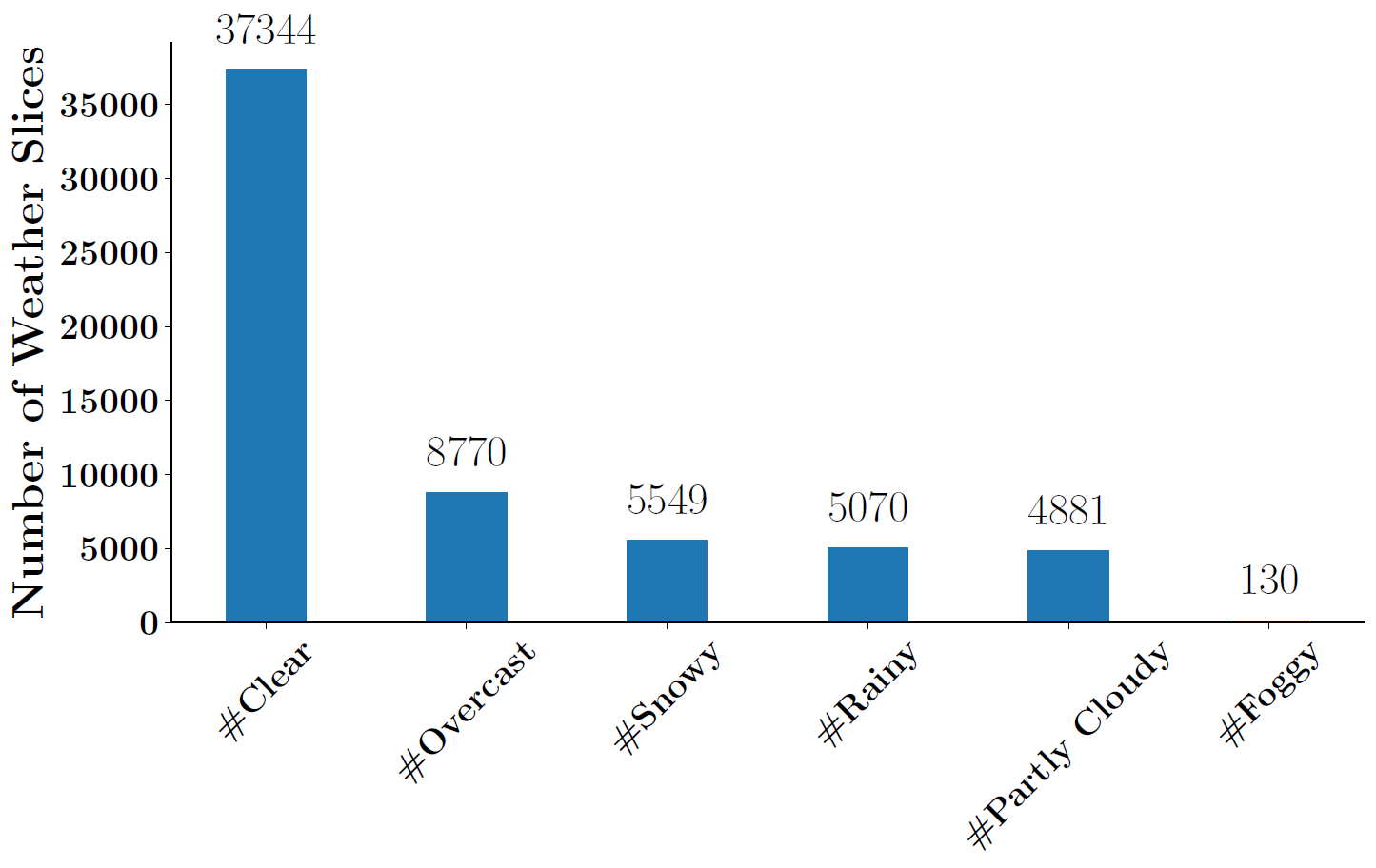}
    \caption{A real-world example of slice-level imbalance in BDD100K~\cite{yu2020bdd100k}. Rainy and foggy weather slices are rarer than the clear weather slice.}
    \label{fig:slice_Histogram}
\end{wrapfigure}
tend to be rarer than clear daytime weather scenarios. We illustrate this setting in \figref{fig:Intro} and provide real-world evidence of this scenario-based slice imbalance in \figref{fig:slice_Histogram}, which details the distribution of weather-based slices in the Berkeley DeepDrive (BDD-100K)~\cite{yu2020bdd100k} dataset. 

To generate robust models, the data curated from the data stream needs to be representative of each slice in the task, which presents a challenge for extracting the most out of each slice for active learning selection in this streaming multi-distributional setting. Importantly, ensuring that this is the case offers protection against rare yet critical slices of the data that may be encountered, such as severe rainy or foggy weather conditions that would otherwise be disastrous for a brittle detector. Worryingly, many streaming algorithms and active learning algorithms do not place explicit focus on improving the performance for these rare slices. Motivated by such scenarios, we address the following question in this work: \textit{Can we automatically improve the performance of a machine learning model on rare slices using multi-distributional streaming data?}

\subsection{Related Work} \label{sec:rel_work}

The setting that we study in this work is at the intersection of many disciplines within machine learning, particularly active learning and stream-based learning. We sketch these related areas and outline how our problem domain is distinct from them. 

\noindent \textbf{Active learning:} Much of the Active Learning (AL) literature focuses on concepts of uncertainty and diversity in selecting instances for labeling. For instance, entropy sampling~\cite{alsurvey} selects instances whose predicted class probability distribution has maximal entropy. This simple strategy is surprisingly competitive in low data-redundancy settings~\cite{eval}. BADGE~\cite{badge}, a more recent state-of-the-art approach, selects instances by performing a \textsc{k-means++} initialization~\cite{kmeansplusplus} on each instance's hypothesized gradient embedding vector formed at the model's final fully-connected layer, which encourages selection of both diverse and uncertain instances. Unlike typical AL settings, the streaming setting studied here limits the access to all encountered unlabeled data on a per-episode basis. Consequently, our AL approach aims to select representative instances from the unlabeled stream, capitalizing on occurrences of rare slices. Other works in AL have tackled similar problems in streaming settings, such as the concept drift issue studied in~\cite{driftstreamactive} and the verification latency issue studied in~\cite{streamlatencyactive}; however, our focus is placed on rare data slice performance to effectively handle multiple distributions of data via a streaming selection setting. Other recent work focuses on targeting instances that are semantically similar to a query set and on avoiding instances that are semantically similar to a private set. PRISM~\cite{prism} formulates submodular information measures~\cite{sims,iyer2021generalized} for such guided subset selection across domains and introduces many of the instantiations used in this work. SIMILAR~\cite{similar} applies these ideas for AL selection in realistic scenarios like rare-classes, privacy-preserving AL selection, and their combination. TALISMAN~\cite{kothawade2022talisman} studies targeted AL for object detection in an offline setting, where all data is available at once. In this paper, we study a more realistic setting where episodes of data are incoming at varying frequencies. 

\textbf{Stream-based learning:}
Closely related to our setting is that of stream-based learning algorithms \cite{fujii2016budgeted,werner2022online,hayes2019memory,hayes2020remind}. Of particular interest is \cite{fujii2016budgeted}, who propose the \textsc{AdaptiveStream} method that uses submodular functions for sampling from an entire stream. Specifically, they partition the entire stream into $k$ partitions and select an item with the maximum marginal gain from each partition, which closely follows the episodic selection of our \textsc{Submodular} baseline in \secref{sec:results}. In another work, \cite{werner2022online} focus on selecting data points in an online setting. They compute a marginal gain for every new incoming data point and propose using a dynamic threshold for filtering the new data point based on the marginal gain. Contrary to these works, our main focus is to leverage a multi-distributional streaming setting with rare yet critical slices. In proximity, both~\cite{hayes2019memory} and~\cite{hayes2020remind} incorporate replay-based incremental/streaming learning algorithms to prevent catastrophic forgetting of past slices; however, our setting differs by allowing for a labeled set that expands by filtering worthwhile unlabeled instances from a buffered high-volume unlabeled stream. We consider this setting due to its high practical utility and, to our knowledge, its novelty in relation to previous works.

\subsection{Our Contributions} \label{sec:contributions}
\begin{wrapfigure}{r}{0.40\textwidth}
    \centering
    \includegraphics[width=0.95\linewidth]{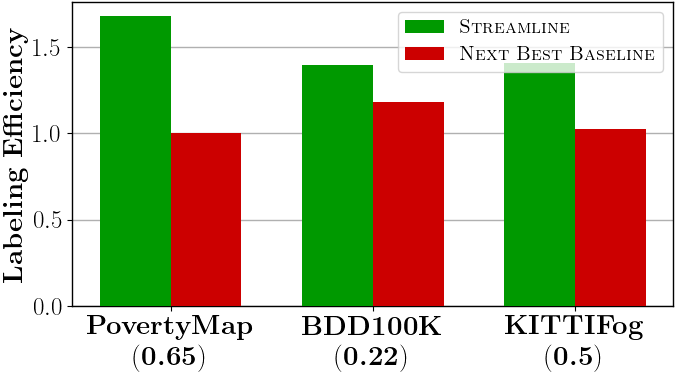}     
    \caption{Labeling efficiencies of \model\ over the next-best performing method on representative datasets. The labeling efficiency shown here depicts how many fewer labels are needed by \model\ and the other method to reach a desired test accuracy (PovertyMap: $65\%$) or test mAP (BDD100K: $22\%$, KITTIFog: $50\%$) when compared to random sampling. The desired values chosen here are the final scores achieved by random sampling on each dataset's rare slice (see \Appref{app:efficiency} for more details).}
    \label{fig:intro_bar}
\end{wrapfigure}
\textbf{\model\ Framework: } In this paper, we propose \model, a novel streaming-based active learning method which follows a unique three-step process to sample from an incoming unlabeled stream. The \textit{first step} identifies the slice category utilizing a normalized submodular mutual information measure~\cite{iyer2021generalized}. The \textit{second step} distributes the labeling budget in an intelligent manner to target rare slices while being more relaxed on common slices. The \textit{third step} selects a subset from the unlabeled stream that is semantically dissimilar to the identified slice's existing labeled data by maximizing the submodular conditional gain~\cite{iyer2021generalized} (see \figref{fig:Streamline_arch}). \looseness-1

\textbf{Effectiveness of \model: } We demonstrate the effectiveness of \model\ over existing AL methods on a wide range of real-world streaming scenarios for a diverse set of datasets across image classification and object detection tasks: iWildCam~\cite{beery2021iwildcam}, PovertyMap~\cite{yeh2020using}, BDD100K~\cite{yu2020bdd100k}, Cityscapes~\cite{cordts2016cityscapes}, and KITTI~\cite{Geiger2012CVPR}. Briefly, we present a representative sample of our results in \figref{fig:intro_bar}. As shown, \model\ requires nearly $1.7\times$ fewer labels than random sampling to reach the same rare slice test accuracy on PovertyMap and roughly $1.4\times$ fewer labels than random sampling to reach the same rare slice test mAP on BDD100K and KITTI, notably improving over the next-best baseline. Overall, we present our full results analysis in \secref{sec:results} and \Appref{app:efficiency}, \ref{app:full_data}, and \ref{app:slice_dist}, where we observe that \model\ outperforms the evaluated existing active learning and stream sampling techniques by up to $5$\% in terms of the rare slice accuracy on our image classification tasks and by up to $8$\% in terms of the rare slice mAP on our object detection tasks.

\textbf{Distinction from \textsc{Similar} and \textsc{Talisman}:}
While \textsc{Similar}~\cite{similar} and \textsc{Talisman}~\cite{kothawade2022talisman} both perform targeted selection for rare-class scenarios, both tend to be unequipped for episodic streaming settings. Indeed, both methods 1) are not adaptive in the manner by which they perform guided selection; 2) oversample common slices of the data by using a fixed budget during each AL selection; and 3) may repeatedly choose redundant data in subsequent selection rounds when performing targeted selection. Each of these facets hampers performance in episodic streaming settings, where 1) the targeted rare slice is not present in the stream, 2) common slices appear much more frequently than rare slices, and 3) the data is locally redundant in time (e.g., frames from a video stream). Importantly, we show that \textsc{Similar}'s targeted selection achieves suboptimal performance in the episodic streaming setting in~\secref{sec:results}, performing consistently worse than \model\ and select baselines. To overcome these challenges, \model\ incorporates selection mechanisms to more effectively target novel rare-slice instances and apportion the labeling budget. This is achieved by 1) identifying the slice within the stream, 2) calculating a fairly apportioned labeling budget, and 3) selecting semantically novel rare-slice instances. We describe each component in more detail in~\secref{sec:results} and show promising results in~\secref{sec:results}.

\section{Preliminaries}
\label{sec:preliminaries}

Here, we present some preliminary notation and concepts that will be used in the development of \model\ and the associated baselines. 

\noindent \textbf{Submodular Functions:}
We draw upon the modeling power of submodular functions and the recently proposed submodular information measures~\cite{sims}. Let $\mathcal{V}$ denote a ground set of data instances. We define a set function $F:2^\mathcal{V}\rightarrow \mathbb{R}$ that assigns scores to subsets of $\mathcal{V}$. Hence, an optimization problem is defined around the maximization of $F$ to find an optimal subset of constrained size with respect to the desired score. However, exactly optimizing $F$ is often intractable since there are $2^{|\mathcal{V}|}$ possible subsets to consider. However, if $F$ has the monotone and submodular properties of set functions, then a simple $(1-\frac{1}{e})$-approximate greedy algorithm finds such a subset~\cite{nemhauser} (see Algorithm~\ref{alg:submod} in Appendix~\ref{app:submod}). $F$ is said to be \emph{submodular} if it has the diminishing returns property: $F(A\cup \{b\}) - F(A) \geq F(B\cup \{b\}) - F(B), A \subseteq B, b \notin B$. $F$ is also said to be \emph{monotone} if $F(A \cup \{b\}) \geq F(A), b \notin A$.

\noindent \textbf{Submodular Information Measures:} Monotone submodular maximization has been used to solve a myriad of combinatorial optimization problems. Interestingly, submodularity has also been used to generalize common information measures, such as Shannon's entropy.~\cite{sims} introduce submodular variants of mutual information and conditional gain that inherit the salient properties of submodular functions.~\cite{sims} define \emph{submodular mutual information} (\textbf{SMI}) as $I_F(A;B) = F(A) + F(B) - F(A \cup B)$ and \emph{submodular conditional gain} (\textbf{SCG}) as $H_F(A|B) = F(A \cup B) - F(A)$ for $A,B \subseteq \mathcal{V}$. Furthermore,~\cite{sims} show that each is submodular in $A$ for a fixed $B$ under certain conditions, which allows for monotone submodular maximization of $I_F(A;B)$ and $H_F(A|B)$. This framework provides a powerful method for performing targeted subset selection by maximizing $I_F(A;Q)$ for some query set $Q$ and for performing privacy-preserving subset selection by maximizing $H_F(A|P)$ for some private set $P$ as studied in~\cite{prism}.

\section{Streamline: Our Streaming Active Learning Framework} \label{sec:ourMethod}

\begin{algorithm}
    \caption{\model}
    \label{alg:ours}
    \begin{algorithmic}[1]
        \STATE {\bfseries Input:} Labeled buffer $\Lcal$ with $T$ slice partitions $P_i$, Unlabeled buffer $\Ucal$, Model $M$, Model weights $\theta$, round AL selection budget $B$, Minimum budget fraction $\rho$
        \STATE $\gamma \leftarrow 0$ \textcolor{blue}{\{Accumulated excess budget\}}
        \REPEAT
            \STATE $t \leftarrow$\textsc{SMIdentify}($\Lcal$, $\Ucal$, $M$, $\theta$) \textcolor{blue}{\{\AlgRef{alg:ident}\}}
            \STATE $b,\gamma \leftarrow$ \textsc{SliceAwareBudget}($\Lcal$, $B$, $\rho$, $\gamma$, $t$) \textcolor{blue}{\{\AlgRef{alg:budget}\}}
            \STATE $\Lcal \leftarrow$ \textsc{SCGSelect}($\Lcal$, $\Ucal$, $M$, $\theta$, $b$) \textcolor{blue}{\{\AlgRef{alg:select}\}}
            \STATE $\theta \leftarrow \text{Train}(\Lcal, M, \theta)$
        \UNTIL converged
    \end{algorithmic}
\end{algorithm}

\begin{figure}[t]
\centering
\includegraphics[width = \textwidth]{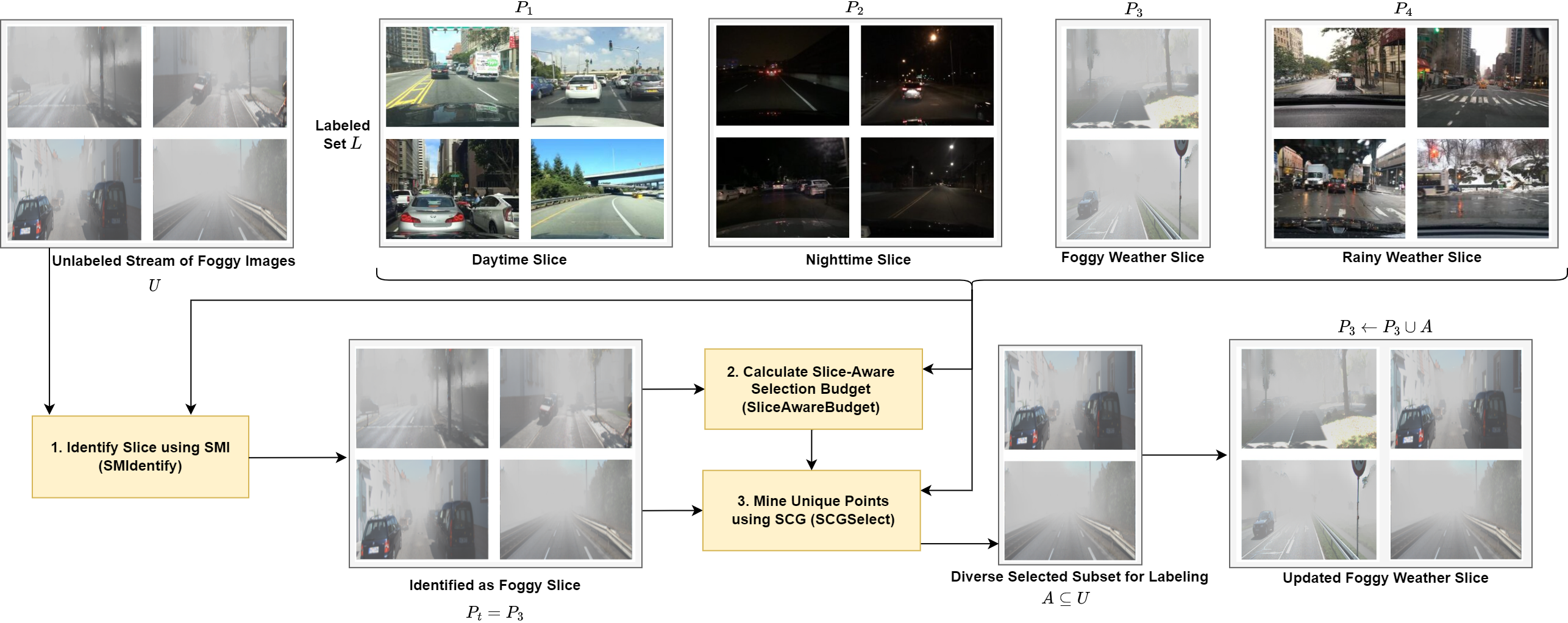}
\caption{Architecture of \model\ for one round of streaming active learning. \textbf{Step 1:} The slice type is identified as `foggy' using \model's \textit{SMI-based identification} scheme. \textbf{Step 2:} \textit{Slice-aware budgeting} assigns a higher budget due to the rarity of the foggy weather slice. \textbf{Step 3:} \model's \textit{SCG-based selection} scheme chooses novel foggy images for labeling, which are subsequently added to the labeled foggy slice.}
\label{fig:Streamline_arch}
\end{figure}

In this section, we propose \model, a streaming active learning framework for realistic multi-distributional settings (see \secref{sec:Intro} for an example). \model\ addresses the limitations of current methods for this setting via a three-step approach using submodular information measures~\cite{iyer2021generalized}. We summarize \model\ in \AlgRef{alg:ours} and \figref{fig:Streamline_arch}. For an arriving episode of unlabeled data in $\Ucal$, we first identify the slice $P_t \subseteq \Lcal$ to which the data in $\Ucal$ belongs using submodular mutual information functions (\textsc{SMIdentify}, Line 4). Based on the identified slice, \model\ automatically allocates a selection budget $b$ based on a budget accumulation scheme that allocates larger $b$ for rare slices and smaller $b$ for common slices (\textsc{SliceAwareBudget}, Line 5). Next, we find a subset of the unlabeled data $A \subseteq \Ucal$ that is semantically dissimilar to $P_t$ using submodular conditional gain functions, which effectively captures a diverse set of representative data to add to the $P_t$ slice of $\Lcal$ (\textsc{SCGSelect}, Line 6). For the remainder of this section, we describe various components and design considerations for each of these three steps. Scalability considerations are discussed in \Appref{app:submod}.

\begin{algorithm}
    \caption{\textsc{SMIdentify}}
    \label{alg:ident}
    \begin{algorithmic}[1]
        \STATE {\bfseries Input:} Labeled buffer $\Lcal$ with $T$ slice partitions $P_i$, Unlabeled buffer $\Ucal$, Model $M$, Model weights $\theta$
        \STATE Calculate the $T$ $|\Ucal| \times |P_i|$ similarity kernels used by FLQMI via a featurization of $P_i$ and $\Ucal$
        \STATE $t \leftarrow \argmax_i \frac{I_F(\Ucal; P_i)}{N_F(\Ucal; P_i)}$ 
        \STATE \textbf{return} $t$
    \end{algorithmic}
\end{algorithm}
\noindent \textbf{Step 1: Slice Identification with SMI (\textsc{SMIdentify}):} To identify the slice $P_t$ of $\Lcal = \bigcup_i P_i$ to which the arriving episode of data in $\Ucal$ belongs, we make use of submodular mutual information~\cite{sims} to find the $P_t$ that is most semantically similar to $\Ucal$, which indicates that $\Ucal$ belongs to slice $P_t$. Specifically, we evaluate the Facility Location Variant Mutual Information (FLQMI) function~\cite{prism} to measure semantic similarity: $I_F(\Ucal;P_t) = \sum_{i \in \Ucal} \max_{j \in P_t} S_{ij} + \sum_{i \in P_t} \max_{j \in \Ucal} S_{ij}$, where $S$ denotes the similarity kernel between elements of $P_t$ and $\Ucal$. By enumerating over each slice partition $P_i \subseteq \Lcal$, \model\ determines the most semantically similar slice by selecting the maximum $I_F(\Ucal;P_i)$ value. As an additional measure, \model\ normalizes $I_F(\Ucal;P_i)$ by a normalizing function $N_F(\Ucal;P_i) = |\Ucal| + |P_i|$ to account for FLQMI's tendency to scale with the size of $P_i$, which allows \model\ to distinguish between slices of different size. In summary, the identification procedure used by \model\ is detailed in \AlgRef{alg:ident}.

\begin{algorithm}
    \caption{\textsc{SliceAwareBudget}}
    \label{alg:budget}
    \begin{algorithmic}[1]
        \STATE {\bfseries Input:} Labeled buffer $\Lcal$ with $T$ slice partitions $P_i$, round AL selection budget $B$, Minimum budget fraction $\rho$, Saved budget $\gamma$, Task identity $t$
        \IF{$P_t$ is a rare slice}
            \STATE Compute $d$, the difference between $|P_t|$ and the average of common $|P_i|$  
            \STATE $\sigma \leftarrow \max (\min (\gamma, B - d), 0)$
            \STATE $b \leftarrow B + \sigma; \gamma \leftarrow \gamma - \sigma$
        \ELSE
            \STATE $\beta \leftarrow \min_i |P_i|$
            \STATE $b \leftarrow B\rho + (1-\rho)B\frac{\beta}{|P_t|}$
            \STATE $\gamma \leftarrow \gamma + (B - b)$
        \ENDIF
        \STATE \textbf{return} $b,\gamma$
    \end{algorithmic}
\end{algorithm}
\noindent \textbf{Step 2: Slice Aware Budgeting (\textsc{SliceAwareBudget} ):}
While traditional pool-based methods select a fixed amount of unlabeled data to label (for example, BADGE~\cite{badge}), this aspect of these strategies does not translate effectively to the episodic streaming setting studied here, where AL methods can benefit greatly on rare slices by apportioning their budget according to the type of data arriving in $\Ucal$. As such, the second step of \model\ uses the calculated slice identity of the previous step to intelligently allocate more of the AL budget to rare slices while reserving the AL budget on common slices. The budgeting scheme is detailed in \AlgRef{alg:budget}. Briefly, \textsc{SliceAwareBudget} calculates the selection budget by incorporating a base selection budget $B$ and a saved budget parameter $\gamma$ that accumulates excess budget when encountering common slices and dissipates this excess when encountering rare slices. For common slices, the final selection budget $b$ is calculated by reserving a minimum fraction of the base budget $\rho B$ and down-scaling the remaining fraction $(1-\rho)B$ using the size of the identified slice. The excess budget $B - b$ is accumulated in $\gamma$. For rare slices, some or all of the saved budget $\gamma$ is used to bring the size of the identified rare slice to the average size of the common slices. This is achieved by transferring this portion of $\gamma$ to $b$. Hence, \model\ effectively capitalizes its selection for choosing rare-slice data from $\Ucal$ via its budgeting scheme in \AlgRef{alg:budget}.

\begin{algorithm}
    \caption{\textsc{SCGSelect}}
    \label{alg:select}
    \begin{algorithmic}[1]
        \STATE {\bfseries Input:} Labeled buffer $\Lcal$ with $T$ slice partitions $P_i$, Unlabeled buffer $\Ucal$, Model $M$, Model weights $\theta$, Selection budget $b$
        \STATE Calculate the $|\Ucal| \times |\Ucal \cup P_t|$ similarity kernel used by FLCG via a featurization of $P_t$ and $\Ucal$. 
        \STATE $X \leftarrow \argmax_{A\subseteq \Ucal,|A|\leq b} H_F(A|P_t)$
        \STATE Label $X$ and augment $P_t \leftarrow P_t \cup X$
        \STATE \textbf{return} $\Lcal$
    \end{algorithmic}
\end{algorithm}
\textbf{Step 3: Selecting Dissimilar Samples (\textsc{SCGSelect}):}
After calculating the selection budget and the slice identity of $\Ucal$, \model\ then performs its AL selection by using submodular conditional gain functions as acquisition functions (e.g.,~\cite{similar}) to select data points that add novel information to the slice $P_t$ (\AlgRef{alg:select}). In many streaming environments, the data collected is often locally redundant, which presents an issue for discerning the worthwhile instances to select from $\Ucal$ (frames from an autonomous vehicle's front camera, for example, tend to be repetitive). SCG functions have been shown to select novel instances in~\cite{similar}; hence, we propose using these to mine a diverse set of novel instances to add to the identified labeled slice. \model\ specifically utilizes the Facility Location Conditional Gain (FLCG) function~\cite{prism, iyer2021generalized}, $H_F(A|P_t)=\sum_{i \in \mathcal{U}} \max \left( \max_{j \in A} S_{ij} - \max_{j \in P_t} S_{ij}, 0 \right)$, where $S$ denotes two similarity kernels: one between elements of $\Ucal$ and $\Ucal$ (first term) and one between elements of $P_t$ and $\Ucal$ (second term). Once instantiated, \model\ uses monotone submodular maximization (see \secref{sec:preliminaries}) to select the subset of novel instances to add to $P_t$. The selection procedure of \model\ is summarized in \AlgRef{alg:select}.

\begin{wrapfigure}{r}{0.4\linewidth}
    \centering
    \includegraphics[width=0.95\linewidth]{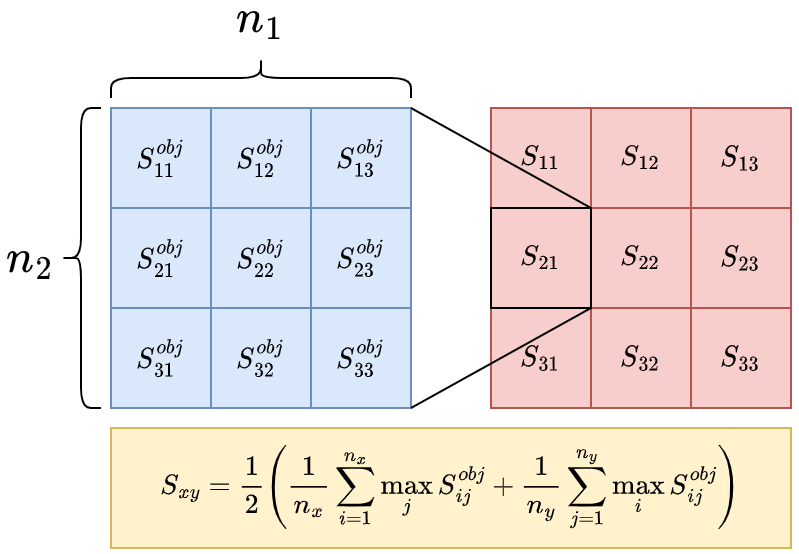}
    \caption{Reduction of object-to-object similarities into a final image-to-image similarity.}
    \label{fig:kernel}
\end{wrapfigure}

\noindent \textbf{Similarity Kernels for Image Classification and Object Detection:}
For both identification and selection, the submodular information measures used by \model\ require similarity kernels to be instantiated. To do so, similarity scores are computed between elements of the labeled data and the unlabeled data by 1) representing each element in an intermediate space and 2) by calculating similarity between these representations. It is common to use the CNN-extracted features and cosine/RBF similarity. However, object detection models can additionally produce multiple object-to-object feature vectors \emph{per image}, potentially yielding more informative measures of similarity between objects. To produce more informative image-to-image similarity kernels for \model's use in object detection, we reduce the object-to-object similarities between two images into one score, as shown by \figref{fig:kernel}. Intuitively, the first summation captures how well the first instance's objects are covered by the second instance's objects. 
Conversely, the second summation captures how well the second instance's objects are covered by the first instance's objects. By computing the average of averages, \model\ captures the similarity between images $x_1$ and $x_2$ based on their object-to-object similarities.

For feature representations of each data element in our image classification study, \model\ utilizes the last linear layer embeddings before computing the pairwise similarity between instances. For our object detection study, \model\ utilizes two separate featurizations of the data for identifying the slice and selecting instances. For identifying, the backbone features of a pre-trained PSPNet semantic segmentation model~\cite{zhao2017pyramid} are used to elicit representations that encode information about the scene, which is better suited for slice-based identification purposes. For selecting, the object-level features obtained from the object detection model being trained are used when computing similarity scores (described above), which better conveys object-to-object similarities that are important for detection performance.

\section{Experimental Results} \label{sec:results}
We evaluate the effectiveness of \model\ on multiple real-world datasets with multi-distributional streaming data, showing improvement over baseline methods in both image classification and object detection settings. Here, we detail each baseline method used in each setting, along with setting formulations for the 5 multi-distributional datasets that we examine: iWildCam~\cite{beery2021iwildcam}, PovertyMap~\cite{yeh2020using}, BDD-100K~\cite{yu2020bdd100k}, Cityscapes~\cite{cordts2016cityscapes}, and KITTI~\cite{Geiger2012CVPR}. Additional details and analysis (including performance on full data) are provided in \Appref{app:efficiency}, \ref{app:full_data}, \ref{app:slice_dist}, and \ref{app:exp_details}.

\subsection{Image Classification} \label{subsec:imgClassification}

\noindent \textbf{Baseline Methods:}
For our image classification experiments, we utilize five baseline methods for comparison against \model: \textsc{Badge}~\cite{badge}, \textsc{Entropy}~\cite{alsurvey}, \textsc{Submodular}~\cite{fujii2016budgeted}, \textsc{Similar}~\cite{similar}, and \textsc{Random}. \textsc{Badge}~\cite{badge} selects a diverse batch of uncertain instances as previously detailed in \secref{sec:Intro}. 
\textsc{Entropy}~\cite{alsurvey} selects the top $k$ instances from the unlabeled data that have the highest entropy over the predicted class probability scores. \textsc{Submodular} selects $k$ instances by finding a subset $A$ that maximizes the Facility Location (FL) function $F(A) = \sum_{i \in \Ucal} \max_{j \in A} S_{ij}$. \textsc{Similar}~\cite{similar} selects $k$ instances by finding a subset $A$ that maximizes the FLQMI function $I_F(A;P_t) = \sum_{i \in P_t} \max_{j\in A} S_{ji} + \sum_{i \in A} \max_{j \in P_t} S_{ij}$. We provide more details regarding the submodular baseline and \textsc{Similar} in~\Appref{app:exp_details}.

\noindent \textbf{Datasets:}
To determine the performance of \model\ on realistic image classification tasks, we utilize the iWildCam dataset~\cite{beery2021iwildcam} and the PovertyMap dataset~\cite{yeh2020using} -- both made available through the Stanford WILDS benchmark~\cite{koh2021wilds}. In the iWildCam dataset, images of wildlife are classified into one of 182 different animal species. Notably, slices of the data are generated based on the location of different camera trap locations, which all produce images of animals to be used in a classification setup. For our experiments, we group all camera trap locations into four slices of consideration and make one of these slices appear infrequently, thereby creating a rare slice. In the PovertyMap dataset~\cite{yeh2020using}, satellite imaging of different regions are used to predict a wealth index score to assist in humanitarian efforts. Here, slices of the data are generated based on whether the images are of rural areas or urban areas. For our experiments, we bin the real-valued index labels to form a binary classification setup and make the urban areas appear infrequently, thereby creating a rare slice.

\noindent \textbf{Setup and Results:}
Using these baselines and datasets, we formulate an initial dataset that has a $5:1$ imbalance between the size of the common slices and the size of the rare slice. We then train a DenseNet-161 model~\cite{densenet} using SGD with cross-entropy loss and random flip augmentations ($p=0.5$) for a maximum of 500 epochs or until training accuracy converges to $99\%$. For each AL selection, we impose a per-round budget of $500$ instances for PovertyMap and $250$ instances for iWildCam. To formulate the streaming component, we replace the unlabeled data after each AL round, presenting the rare slice every 3 rounds in PovertyMap and presenting each slice sequentially in iWildCam. We repeat these experiments for a total of 4 runs to produce error bars. For \model\, we utilize a minimum budget fraction of $\rho = 0.825$ for PovertyMap and $\rho = 0.5$ for iWildCam.

\begin{figure}[t]
    \centering
    \includegraphics[width=0.95\textwidth]{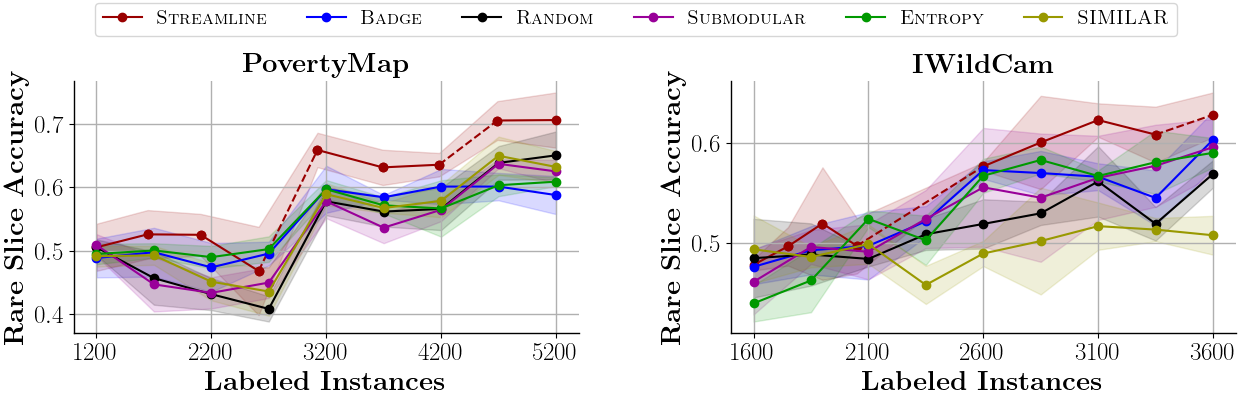}
    \caption{Performance of \model\ versus baseline methods on image classification datasets from the WILDS benchmark~\cite{koh2021wilds}. The dashed line for \model\ indicates when \model\ adds labeled rare slice data and is spending the accumulated budget.  Notably, \model\ achieves an improvement of $5$\% in rare slice accuracy (urban settings) on PovertyMap~\cite{yeh2020using} and an improvement of $3$\% in rare slice accuracy (rare camera trap locations) on iWildCam~\cite{beery2021iwildcam}.}
    \label{fig:img_cls}
\end{figure}

The rare-slice accuracy is presented in \figref{fig:img_cls}. In both cases, \model\ achieves the highest rare-slice accuracy on both datasets, superseding the baselines by $3\%$ on iWildCam and $5\%$ on PovertyMap. Notably, \model\ correctly leverages the occurrence of each rare slice, which is determined by the sharp jump in accuracies denoted by a dashed line between the fourth and fifth data points in both plots. This is due to the joint modeling of slice-relevance and representation in the FLQMI function and modeling of dissimilarity with existing labeled data in the FLCG function. \figref{fig:img_cls} additionally shows the budgeting scheme of \model\ in effect with the identification scheme, choosing to sample fewer instances for frequent occurring slices (denoted by solid lines). Hence, \model\ achieves superior performance on rare-slice accuracy by combining the slice identification, budgeting, and SCG-based selection schemes, notably improving over the performance of SIMILAR~\cite{similar} as discussed in~\secref{sec:contributions}.

\subsection{Object Detection} \label{subsec:objDetection}

\noindent \textbf{Baseline Methods:}
For our object detection experiments, we utilize five baseline methods for comparison against \model: \textsc{Entropy}~\cite{alsurvey}, \textsc{Least Conf.}~\cite{alsurvey}, \textsc{Margin}~\cite{alsurvey}, \textsc{Submodular}~\cite{fujii2016budgeted}, and \textsc{Random}. Notably, most models used in object detection generate a number of bounding box predictions, each with 4 coordinates to determine the box and a vector of class probabilities to determine the object within the box. To adapt the common uncertainty methods of \textsc{Entropy}, \textsc{Least Conf.}, and \textsc{Margin} to the object detection setting, we follow the scheme of~\cite{kothawade2022talisman} to consolidate the uncertainty scores for each \emph{object} into a score for each \emph{image} by averaging these scores (see \Appref{app:exp_details}). \textsc{Submodular} follows the same procedure as the image classification experiments; however, the similarity kernel is instantiated using the CNN backbone features of the object detection model. See \Appref{app:exp_details} for more details on the baselines.

\noindent \textbf{Datasets:}
To determine the performance of \model\ on realistic object detection tasks, we utilize three autonomous driving datasets: BDD-100K~\cite{yu2020bdd100k}, Cityscapes~\cite{cordts2016cityscapes}, and KITTI~\cite{Geiger2012CVPR}. In BDD-100K, slices are generated based on time-of-day metadata. Specifically, we formulate two data slices: night images versus day images, where we make the night images appear less frequently. For Cityscapes and KITTI, we generate infrequent rain-occluded and fog-occluded weather slices, respectively, using the physics-based rendering techniques of~\cite{halder2019physics}. Hence, all three datasets used in our experiments feature a common slice and a rare slice whose performance is critical for safe autonomous vehicle performance.

\noindent \textbf{Setup and Results:}
As with our image classification experiments, we formulate an initial dataset that has a $5:1$ imbalance between the size of the common slice and the size of the rare slice. We then train a Faster-RCNN model~\cite{ren2015faster} using SGD with a mixture of cross-entropy loss and L1 loss (for the RPN and RoI heads) and random flip augmentations for 100 epochs via MMDetection~\cite{mmdetection}. For each AL selection, we impose a per-round budget of $250$ instances for BDD-100K and KITTI and $200$ instances for Cityscapes. As before, we replace the unlabeled data after each AL round to formulate the streaming component, presenting the rare slice every 3 rounds in each dataset. To additionally introduce the redundancy of the frames usually present in autonomous vehicle settings, we make the incoming unlabeled data redundant by ensuring each instance has two copies in the unlabeled data. For \model\, we utilize a minimum budget fraction of $\rho = 0.5$ for all datasets.

\begin{figure*}[t]
    \centering
    \includegraphics[width=0.9\textwidth]{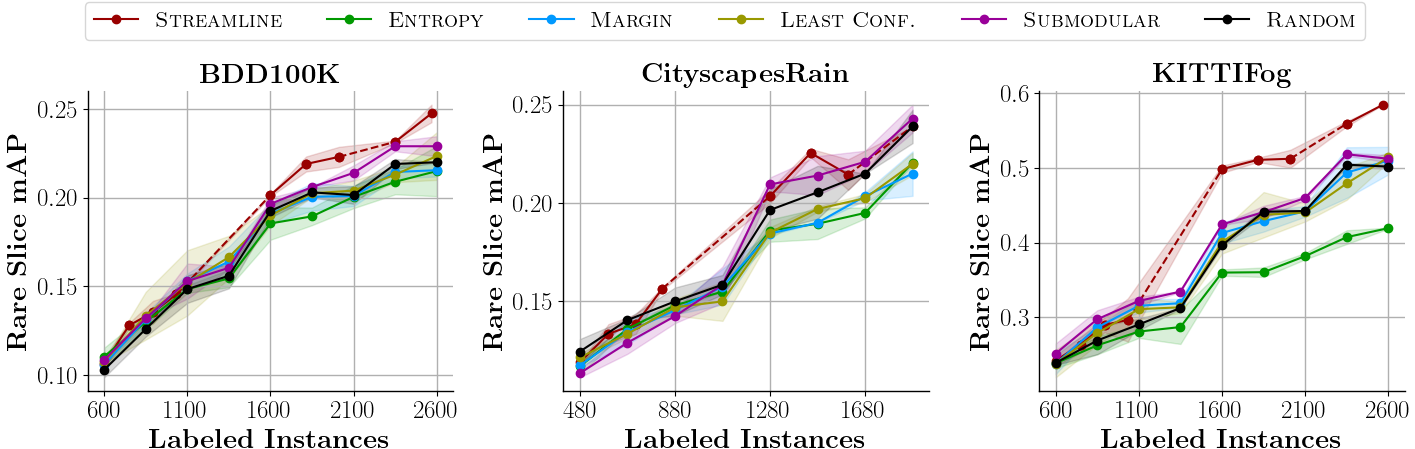}
    \caption{Performance of \model\ versus baseline methods on object detection datasets. The dashed line for \model\ indicates that \model\ has detected a rare slice and is spending the accumulated budget on this slice. \model\ achieves an improvement of $2$\% in rare slice mAP (nighttime settings) on BDD100K~\cite{yu2020bdd100k} and $8$\% in rare slice mAP (foggy settings) on KITTI~\cite{Geiger2012CVPR, halder2019physics}. \model\ remains comparable to the baselines on Cityscapes~\cite{cordts2016cityscapes, halder2019physics}.}
    \label{fig:obj_det}
\end{figure*}

The rare-slice mAP is presented in~\figref{fig:obj_det}. In general, \model\ outperforms the baselines across each dataset. In our Cityscapes rain experiments, \model\ maintains higher rare-slice mAP than the baselines, ending with comparable mAP to the baselines. \model\ achieves a larger mAP advantage over the baselines of $2\%$ in our night \emph{vs.} day BDD-100K experiments. Lastly, \model\ achieves the highest mAP advantage over the baselines of nearly $8\%$ in our KITTI fog experiments. Hence, \model\ has the potential to greatly outperform existing methods when improving rare slice performance. Indeed, \model\ tends to exhibit larger gains in rare-slice mAP depending on the degree to which objects are obscured, ranging from the smallest gain in mAP for the slightly occluded rainy images to the largest gain in mAP for the heavily occluded foggy images. Owing to the joint modeling of query-relevance and representation capability of FLQMI, \model\ correctly identifies the incoming slices. Subsequently, it applies its budgeting scheme effectively to mine unique data points from the rare slice using FLCG (denoted by dashed lines in \figref{fig:obj_det}) and significantly improve the rare-slice mAP. 

\subsection{Ablation}

\begin{wrapfigure}{r}{0.45\linewidth}
    \centering
    \includegraphics[width=0.95\linewidth]{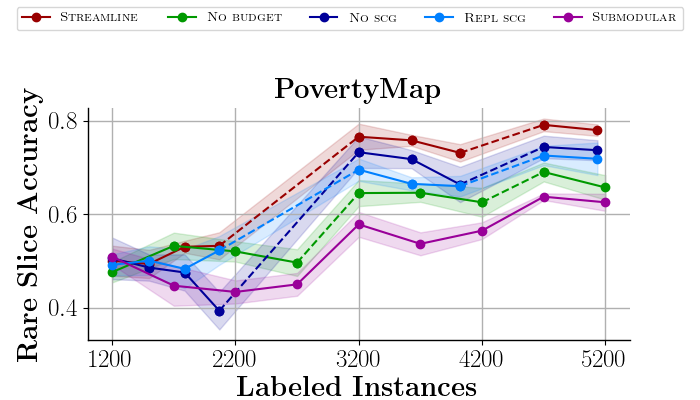}
    \caption{Ablation study of \model\ on the PovertyMap dataset. Each component of \model\ contributes to its effectiveness.}
    \label{fig:ablation}
\end{wrapfigure}
To conclude our analysis, we study the contribution of each of the three main components of \model: \textsc{SMIdentify} (\AlgRef{alg:ident}), \textsc{SliceAwareBudget} (\AlgRef{alg:budget}), and \textsc{SCGSelect} (\AlgRef{alg:select}). We study three variants of \model\ that have one of these components removed. \textsc{No scg} negates the SCG component with \textsc{Random} while \textsc{Repl scg} replaces it with \textsc{Badge}~\cite{badge}, a strong AL method. \textsc{No budget} replaces the budgeting scheme with the fixed budget used by the baselines. The last variant removes the slice identification component; however, without task information, slice-aware budgeting and slice-based conditioning via SCG cannot be performed. Accordingly, we use the \textsc{Submodular} baseline to represent this variant. We show the performance of each variant versus \model\ ($\beta = 0.5$) on PovertyMap~\cite{yeh2020using} in \figref{fig:ablation}. We find that the slice-aware budgeting step contributes the most to \model's performance, followed by the SCG-based selection scheme. Both of these ablated versions outperform the \textsc{Submodular} baseline. 


\section{Conclusion} \label{sec:conclusion}

In this work, we highlight the vulnerability of existing AL methods for the episodic multi-distributional streaming setting present in many realistic scenarios, such as autonomous vehicle data collection and model training. In such settings, model performance on rare slices of data must be nominal to ensure safe application of those models. Motivated by recent advances in submodular information measures~\cite{iyer2021generalized}, we develop a new streaming AL method called \model\ that identifies rare-slice arrivals in the stream, apportions the selection budget for these slices, and performs AL selection that uses slice-based information to improve performance on rare slices of the data. The modular design of \model\ allows for its application in many settings, including image classification and object detection settings. Empirically, we show that \model\ outperforms existing AL methods on realistic datasets in both image classification and object detection settings, solidifying its application in a variety of real-world settings.

\clearpage

\bibliography{main}
\bibliographystyle{unsrt}


\newpage
\appendix

\begin{center}
\part{Supplementary Material for STREAMLINE: Streaming Active Learning for Realistic Multi-Distributional Settings} 
\end{center}
\parttoc 

\section{Submodular Maximization and Scalability Tricks}
\label{app:submod}

\subsection{Submodular Maximization Algorithms}

\begin{algorithm}
    \caption{Cardinality-Constrained Submodular Maximization}
    \label{alg:submod}
    \begin{algorithmic}
        \STATE {\bfseries Input:} Monotone submodular function $F:2^\Vcal \rightarrow \mathbb{R}$ instantiated on ground set $\mathcal{V}$, Cardinality constraint $b$
        \STATE $A \leftarrow \emptyset$
        \FOR{$i=1$ to $b$}
            \STATE $x \leftarrow \text{argmax}_{j \in \mathcal{V} - A} F(A \cup \{j\}) - F(A)$
            \STATE $A \leftarrow A \cup \{x\}$
        \ENDFOR
    \end{algorithmic}
\end{algorithm}

\model\ utilizes monotone submodular maximization heavily in its operation. As discussed previously, a simple greedy algorithm can be used to find a cardinality-constrained subset that maximizes the submodular function to within a $(1-\frac{1}{e})$ factor of the optimal value~\cite{nemhauser}. The simple greedy algorithm is presented here in Algorithm~\ref{alg:submod}. The simple greedy algorithm, then, has a time complexity of $O(bF|\Vcal|)$ (where $F$ is overloaded notation for the function evaluation complexity), which can be restrictive if the size of the ground set is too large or if evaluating $F$ is expensive in practice. To ease this complexity, a variant known as the lazy greedy algorithm (also known as the accelerated greedy algorithm)~\cite{minoux1978accelerated} makes use of a priority queue to evaluate the max. Specifically, noting that the marginal gain of an element decreases with the size of the selected set (via submodularity), the lazy greedy algorithm maintains a priority queue of upper bounds on the marginal gain of each element. When an element is ready to be de-queued from the priority queue, the true marginal gain of that element is evaluated, and its upper bound is updated to this value. If the element remains at the front of the priority queue after this update, then it must be that the element has the highest marginal gain of the elements remaining in the queue via submodularity; subsequently, this element is de-queued as the element to greedily add to the selected subset. This effectively allows the greedy algorithm to "lazily" evaluate $F$ on only a few elements per iteration of the loop in practice, although the worst case complexity remains the same. Importantly, the lazy greedy algorithm maintains the same $(1 - \frac{1}{e})$ approximation factor, which makes it a common choice for submodular maximization.

In some cases, even more scalable submodular solutions than the lazy greedy variant are desirable. To that end,~\cite{mirzasoleiman2015lazier} have proposed another monotone submodular maximization algorithm called the lazier-than-lazy greedy algorithm (also known as stochastic greedy) that loosens the $(1 - \frac{1}{e})$ approximation guarantee to $(1 - \frac{1}{e} - \epsilon)$ in expectation but only requires $O(|\Vcal|\log \frac{1}{\epsilon})$ function evaluations compared to the $O(|\Vcal|b)$ of the simple greedy algorithm. The stochastic greedy algorithm achieves this by only evaluating the max of the greedy algorithm on a randomly chosen subset of $\frac{|\Vcal|}{b}\log \frac{1}{\epsilon}$ elements. Additionally, stochastic greedy can utilize the same priority queue scheme of the lazy greedy algorithm~\cite{mirzasoleiman2015lazier}, greatly accelerating the running time of monotone submodular maximization while empirically matching more classical algorithms in many scenarios. 

\subsection{Partitioning}

Many submodular functions are based on similarity kernels, which tend to scale quadratically with the number of elements in the ground set. For example, the simple Facility Location function (off which FLQMI~\cite{iyer2021generalized, prism} is based) requires a $|\Vcal| \times |\Vcal|$ similarity kernel for its evaluation:

\begin{equation}
    F(A) = \sum_{i \in \Vcal} \max_{j \in A} S_{ij}
\end{equation}

\noindent Hence, Facility Location requires $O(|\Vcal|^2)$ space to be instantiated and requires $O(|\Vcal||A|)$ time to be evaluated. In large-data scenarios, the $|\Vcal|$ factor tends to be very restrictive, especially with regard to space. This limits the applicability of Facility Location and other similarity-based submodular functions. To alleviate the $O(|\Vcal|^2)$ dependency of these functions, many works (such as~\cite{similar}) have resorted to \emph{partitioning} the ground set into $p$ partitions and performing submodular maximization on each with a cardinality constraint of $\frac{b}{p}$. After performing submodular selection on each of the $p$ partitions, the $p$ subsets of size $\frac{b}{p}$ are combined to yield a selected subset of $b$ elements. Hence, the space requirements for these similarity-based submodular functions decreases to $O(\frac{|\Vcal|^2}{p^2})$ per instantiation. If each selection is performed sequentially, the space requirement remains $O(\frac{|\Vcal|^2}{p^2})$ as only one submodular function is instantiated at a time. If each selection is performed in parallel (a time-beneficial aspect of partitioning), then the space requirement increases to $O(\frac{|\Vcal|^2}{p})$. 

As the simple and lazy greedy algorithms require $O(|\Vcal|bF)$ time and the stochastic greedy algorithm requires $O(|\Vcal|F\log \frac{1}{\epsilon})$ time, each experiences the same degree of speedup due to partitioning. In the case of Facility Location, both the lazy greedy and stochastic greedy algorithms respectively scale down to $O(\frac{|\Vcal|}{p}\frac{b}{p}\frac{|\Vcal|}{p}\frac{b}{p})$ and $O(\frac{|\Vcal|}{p}\frac{|\Vcal|}{p}\frac{b}{p}\log \frac{1}{\epsilon})$ (equivalently, $O(\frac{|\Vcal|^2b^2}{p^4}$ and $O(\frac{|\Vcal|^2b}{p^3}\log \frac{1}{\epsilon})$) since the ground set becomes size $\frac{|\Vcal|}{p}$ and $A$ is no larger than $\frac{b}{p}$ in either algorithm. However, both algorithms must be applied for the $p$ partitions, bringing the full running time of \emph{sequentially} maximizing Facility Location with partitioning to $O(\frac{|\Vcal|^2b^2}{p^3})$ using the simple or lazy greedy algorithm and $O(\frac{|\Vcal|^2b}{p^2}\log \frac{1}{\epsilon})$ using the stochastic greedy algorithm. If done \emph{in parallel}, the running time remains $O(\frac{|\Vcal|^2b^2}{p^4})$ using the simple or lazy greedy algorithm and $O(\frac{|\Vcal|^2b}{p^3}\log \frac{1}{\epsilon})$ using the stochastic greedy algorithm.

To summarize, the time and space complexities of partitioning are given in \tabref{tab:partitiontime} and \tabref{tab:partitionspace}, respectively. Overloading notation, $F_{part}$ denotes the running time complexity of $F$ on a partition of the data and $F$ denotes the running time complexity of $F$ on the full ground set. $F_{part}$ is used here since other similarity-based submodular functions have different time complexities than Facility Location. In general, partitioning provides a way to reduce both the space and time complexity of submodular maximization while still retaining the expressive power of submodular functions for reasonable choices of $p$.

\begin{table}[]
    \centering
    \caption{Space complexities of submodular maximization algorithms on similarity-based submodular functions with and without partitioning. Here, both sequential versions and parallel versions of the partitioning scheme are provided.}
    \begin{tabular}{|c|c|c|c|}
        \hline
        & Space & Part. Space (Seq.) & Part. Space (Par.) \\
        \hline
        Simple & $O(|\Vcal|^2)$ & $O(\frac{|\Vcal|^2}{p^2})$ & $O(\frac{|\Vcal|^2}{p})$ \\
        \hline
        Lazy & $O(|\Vcal|^2)$ & $O(\frac{|\Vcal|^2}{p^2})$ & $O(\frac{|\Vcal|^2}{p})$ \\
        \hline
        Stochastic & $O(|\Vcal|^2)$ & $O(\frac{|\Vcal|^2}{p^2})$ & $O(\frac{|\Vcal|^2}{p})$ \\
        \hline
    \end{tabular}
    \label{tab:partitionspace}
\end{table}

\begin{table}[]
    \centering
    \caption{Time complexities of submodular maximization algorithms on similarity-based submodular functions with and without partitioning. Here, both sequential versions and parallel versions of the partitioning scheme are provided.}
    \begin{tabular}{|c|c|c|c|}
        \hline
        & Time & Part. Time (Seq.) & Part. Time (Par.)   \\
        \hline
        Simple & $O(|\Vcal|bF)$ & $O(\frac{|\Vcal|bF_{part}}{p})$ & $O(\frac{|\Vcal|bF_{part}}{p^2})$ \\
        \hline
        Lazy & $O(|\Vcal|bF)$ & $O(\frac{|\Vcal|bF_{part}}{p})$ & $O(\frac{|\Vcal|bF_{part}}{p^2})$ \\
        \hline
        Stochastic & $O(|\Vcal|F\log\frac{1}{\epsilon})$ & $O(|\Vcal|F_{part}\log \frac{1}{\epsilon})$ & $O(\frac{|\Vcal|F_{part}}{p}\log \frac{1}{\epsilon})$ \\
        \hline
    \end{tabular}
    \label{tab:partitiontime}
\end{table}

\subsection{Application to \model}

The subset selection complexity of \model\ is dominated by: i) \textit{slice identification} using SMI functions and ii) \textit{mining unique samples} using SCG functions. For slice identification, a total of $T$ $\Ucal \times P_i$ kernels are computed, resulting in $|\Ucal||\Lcal|$ similarity computations. Additionally, evaluating $I_F(\Ucal;P_i)$ for each $P_i$ requires $2|\Ucal||\Lcal|$ operations, giving a time complexity of $O(|\Ucal||\Lcal|)$. For mining unique samples, we use the FLCG function, which requires a $|\Ucal| \times |\Ucal|$ kernel in addition to a $|\Ucal| \times |P_t|$ kernel. For optimization in our experiments, we use the lazy greedy algorithm~\cite{minoux1978accelerated} to achieve the optimal $(1 - \frac{1}{e})$ approximation guarantee, which allows a worst-case time complexity of $O(|\Ucal|^2 + |\Ucal||P_t| + B^2|\Ucal|^2)$ for instantiating (first two terms) and optimizing (last term) the FLCG function. In total, the full space complexity of \model\ is $O(|\Ucal|^2 + |\Ucal||P_{max}|)$, where $P_{max}$ is the largest labeled slice from the identification step. For streaming settings, the size of $|\Ucal|$ usually remains small since $|\Ucal|$ is periodically refreshed with new data incoming in the stream, which will allow the use of the lazy greedy algorithm in most cases. In scenarios where the volume of data being generated by the stream is particularly large, the scalability techniques discussed here can be applied as is commonly done in submodular optimization to reduce these complexities even further.

Since \model\ utilizes monotone submodular maximization when optimizing the FLCG function (see \secref{sec:ourMethod}), the partitioning trick can be used in our setting. From before, the full running time of the selection step is $O(|\Ucal|^2 + |\Ucal||P_t| + B^2|\Ucal|^2)$, requiring $O(|\Ucal|^2 + |\Ucal||P_t|)$ space (note that $P_{max}$ is exchanged with $P_t$ since $P_{max}$ comes from the identification step). When partitioning, the space requirement reduces to $O(\frac{|\Ucal|^2}{p^2} + \frac{|\Ucal||P_t|}{p})$ for computing the similarity kernels in a sequential manner. If this is done in parallel, the space requirement becomes $O(\frac{|\Ucal|^2}{p} + |\Ucal||P_t|)$. For the time complexity, computing both kernels across partitions sequentially reduces to $O(\frac{|\Ucal|^2}{p} + |\Ucal||P_t|)$ time while computing both kernels across partitions in parallel reduces to $O(\frac{|\Ucal|^2}{p^2} + \frac{|\Ucal||P_t|}{p})$ time. Evaluating the FLCG function with this partitioning requires at most $O(\frac{|\Ucal|B}{p^2})$ time since the sum is over $\frac{|\Ucal|}{p}$ elements, the max over $A$ considers no more than $\frac{B}{p}$ elements (the budget for $A$), and the max over $P_t$ is constant time after first max-reducing $S$ row-wise. Referencing \tabref{tab:partitiontime}, this gives an optimization time of $O(\frac{|\Ucal|^2B^2}{p^3})$ if done sequentially and $O(\frac{|\Ucal|^2B^2}{p^4})$ if done in parallel for the simple and lazy greedy algorithms. Likewise, this gives an optimization time of $O(\frac{|\Ucal|^2B}{p^2}\log \frac{1}{\epsilon})$ if done sequentially and $O(\frac{|\Ucal|^2B}{p^3}\log \frac{1}{\epsilon})$ if done in parallel for the stochastic greedy algorithm. In total, applying partitioning in parallel brings the original $O(|\Ucal|^2 + |\Ucal||P_t| + B^2|\Ucal|^2)$ running time using $O(|\Ucal|^2 + |\Ucal||P_t|)$ space to $O(\frac{|\Ucal|^2}{p^2} + \frac{|\Ucal||P_t|}{p} + \frac{|\Ucal|^2B^2}{p^4})$ using $O(\frac{|\Ucal|^2}{p} + |\Ucal||P_t|)$ space. Applying partitioning sequentially brings the original $O(|\Ucal|^2 + |\Ucal||P_t| + B^2|\Ucal|^2)$ running time using $O(|\Ucal|^2 + |\Ucal||P_t|)$ space to $O(\frac{|\Ucal|^2}{p} + |\Ucal||P_t| + \frac{|\Ucal|^2B^2}{p^3})$ using $O(\frac{|\Ucal|^2}{p^2} + \frac{|\Ucal||P_t|}{p})$ space. Finally, one can apply the stochastic greedy method to get even faster complexities in exchange for a looser optimization guarantee for each partition.

\begin{figure}[t]
    \centering
    \includegraphics[width=0.75\linewidth]{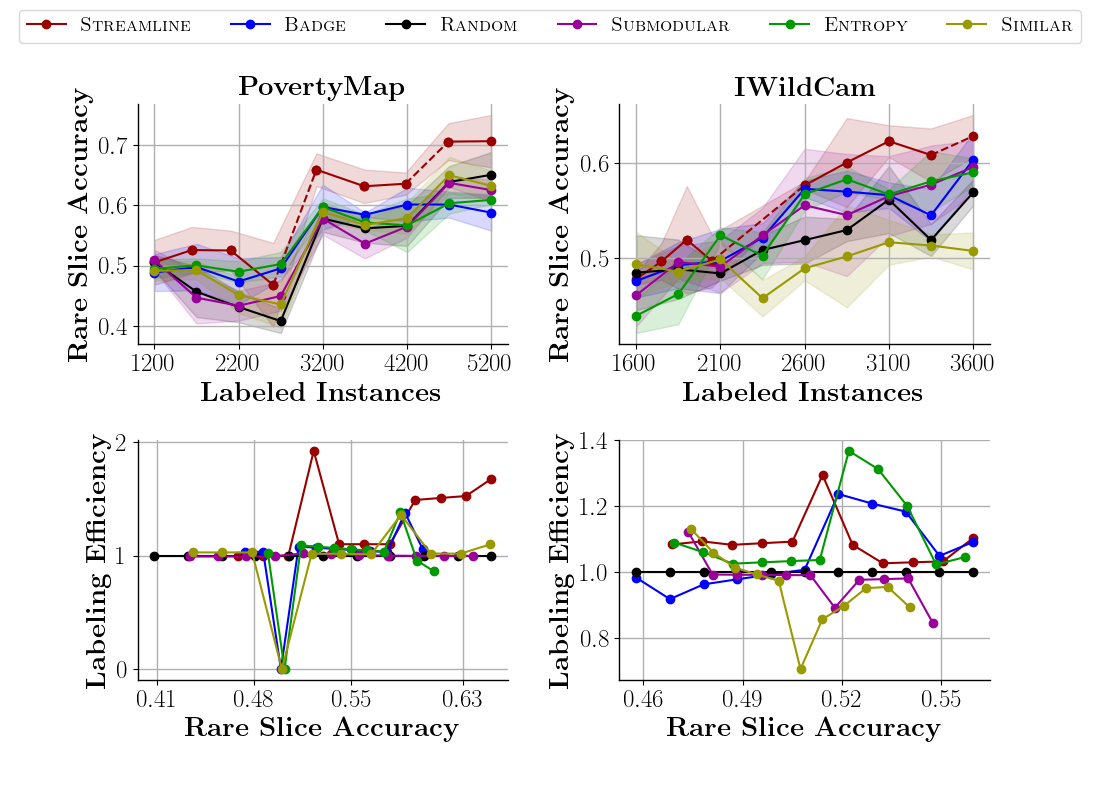}
    \caption{Labeling efficiencies of \model\ and the baseline methods from our image classification experiments. \model\ achieves higher labeling efficiencies in totality across PovertyMap~\cite{yeh2020using} and the initial rounds of iWildCam~\cite{beery2021iwildcam}. For iWildCam, we note that \model\ still achieves the highest rare slice accuracies, and labeling efficiency with respect to random sampling cannot be computed since random sampling does not yield label-accuracy data in that range.}
    \label{fig:img_cls_app}
\end{figure}

As a last scalability consideration for \model, we note that the identification step is not amenable to the optimization algorithms and partitioning tricks discussed here. Furthermore, the complexities of the optimization step have dependencies on the size of the labeled slices. We posit that these factors can be mitigated by devising slice summaries for each labeled slice so that the $P_t$ factors become constant. Such summaries can be obtained via submodular optimization. While these summaries can help improve the scalability of \model, having summaries reduces the amount of information that can be used to more reliably identify the incoming data in the unlabeled stream and select unique instances from the unlabeled stream. As such, we leave this scalability consideration for future study but mention it here to help reduce the dependency on the size of the labeled slices.

\section{Labeling Efficiency of \model\ and Other Baselines}
\label{app:efficiency}

\begin{figure}[h]
    \centering
    \includegraphics[width=0.8\linewidth]{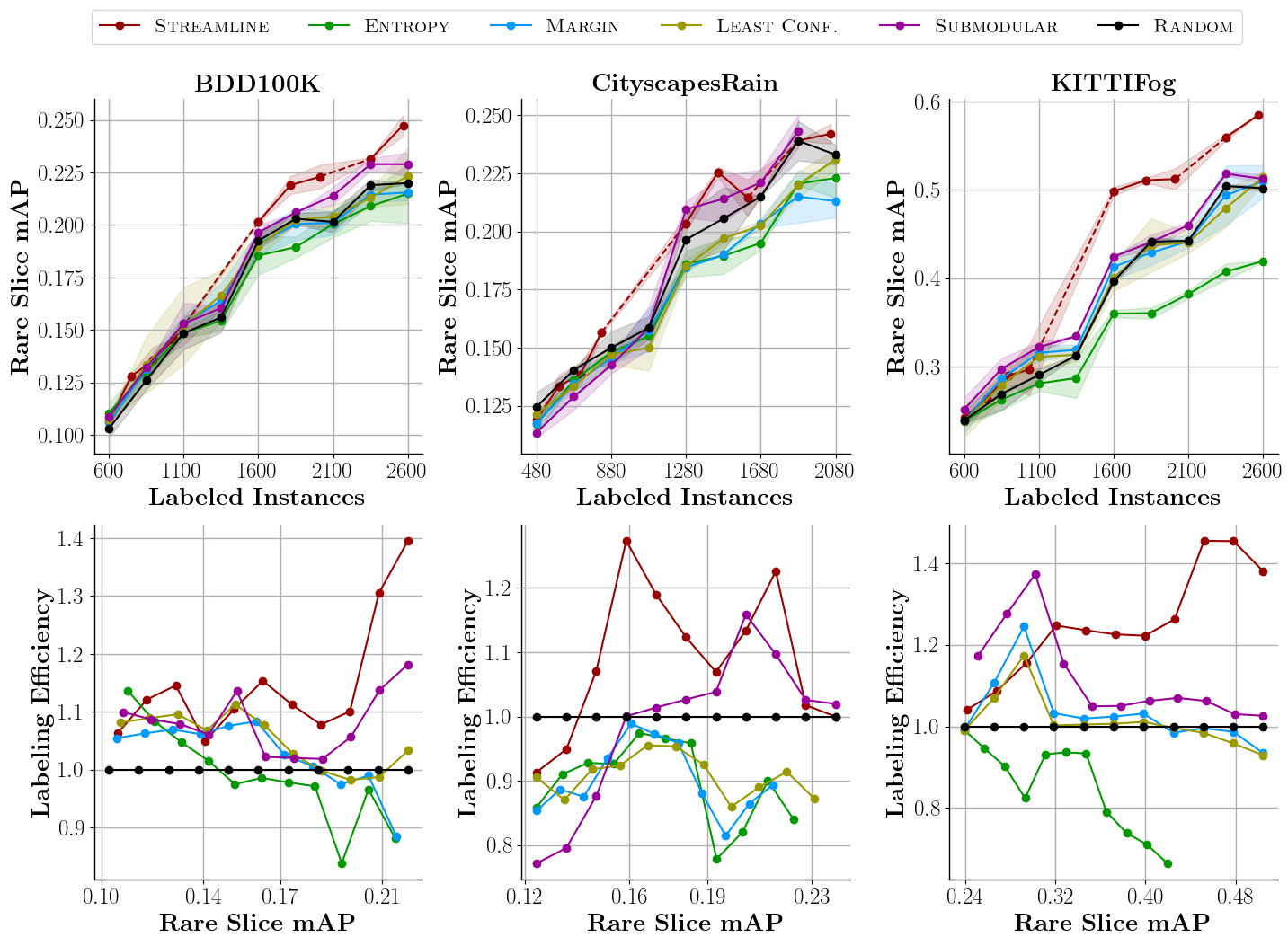}
    \caption{Labeling efficiencies of \model\ and the baseline methods from our object detection experiments. \model\ commandingly achieves higher labeling efficiencies across BDD100K~\cite{yu2020bdd100k}, Cityscapes~\cite{cordts2016cityscapes}, and KITTI~\cite{Geiger2012CVPR}.}
    \label{fig:obj_det_app}
\end{figure}

In this section, we further analyze the results presented in the introduction via \figref{fig:intro_bar} and in \secref{sec:results} under the lens of \emph{labeling efficiency}. While evaluating accuracy/mAP versus the number of labeled instances gives insight into the performance of an active learning method by itself, a crucial aspect about the performance of an active learning method can often be overlooked: the active learning method's ability to yield a target performance with fewer labels versus random sampling. Indeed, random sampling reflects the least-effort strategy present in active learning, and efforts to generate new methods in active learning are often premised on doing substantially better than random sampling. To this end, we calculate the \emph{labeling efficiency} of an active learning method compared to random sampling, which is defined to be the number of labels needed by random sampling to reach a desired test metric versus the number of labels needed by the examined active learning method. For example, if \model\ requires only 100 labels to reach a test metric value of $\alpha$ and random sampling requires 200 labels to reach $\alpha$, then \model\ is said to have a labeling efficiency of $2\times$ with respect to random for test metric value $\alpha$.

Accordingly, this allows us to proceed with further analysis of our experiment results presented in \secref{sec:results}. By determining the labeling efficiency across the shared values of the test metric between random sampling and a compared active learning method, the convergence of the compared active learning method can be better relativized versus random sampling. We present the labeling efficiencies of each method with respect to random sampling in \figref{fig:img_cls_app} and \figref{fig:obj_det_app} for our image classification experiments and object detection experiments, respectively. In our image classification experiments, we see that \model\ enjoys greatly improved labeling efficiency over the baselines in PovertyMap~\cite{yeh2020using}. Across the accuracy ranges of random sampling in our experiments on iWildCam~\cite{beery2021iwildcam}, we see that \model\ enjoys improved labeling efficiency in the lower accuracy ranges but becomes temporarily superseded by the baselines. However, \model\ is shown to outperform the baselines in the higher accuracy ranges that cannot be computed due to a lack of data points from random sampling. In our object detection experiments, we see that \model\ also achieves the highest labeling efficiency across the comparable ranges of mAP, which further distinguishes \model's utility in improving rare slice mAP over the baseline methods. In all experiments, we highlight the sharp increase in labeling efficiency, which are a product of \model's first exposure to the rare slice.

\section{Evaluation on Full Data}
\label{app:full_data}

\begin{figure}[h]
    \centering
    \includegraphics[width=\linewidth]{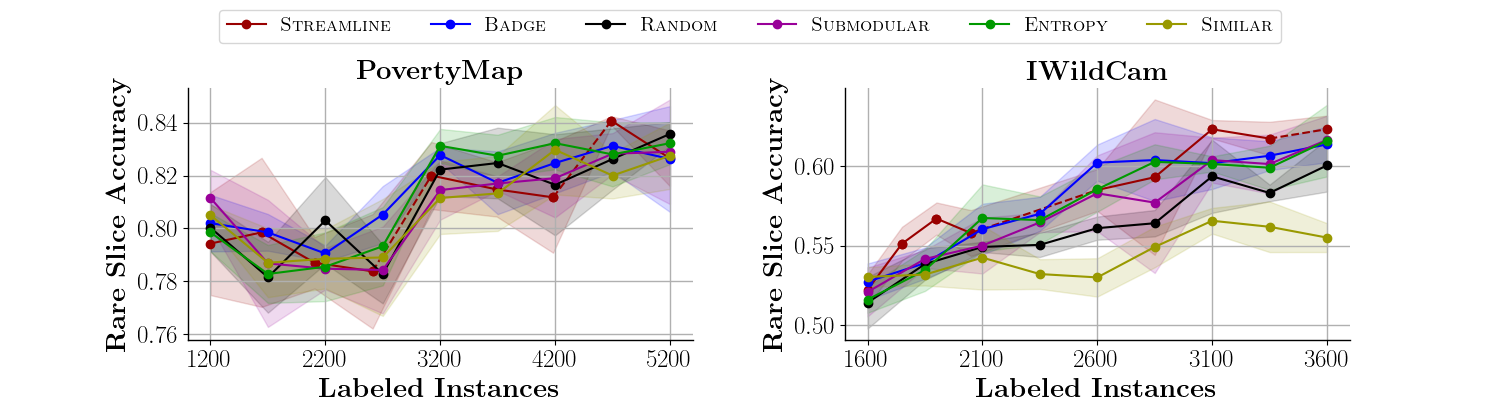}
    \caption{Full accuracy of \model\ and the baseline methods from our image classification experiments. \model\ successfully retains comparable full accuracy to the baselines while achieving higher rare-slice accuracy as shown in~\secref{sec:results}.}
    \label{fig:img_cls_full}
\end{figure}

\begin{figure}[h]
    \centering
    \includegraphics[width=\linewidth]{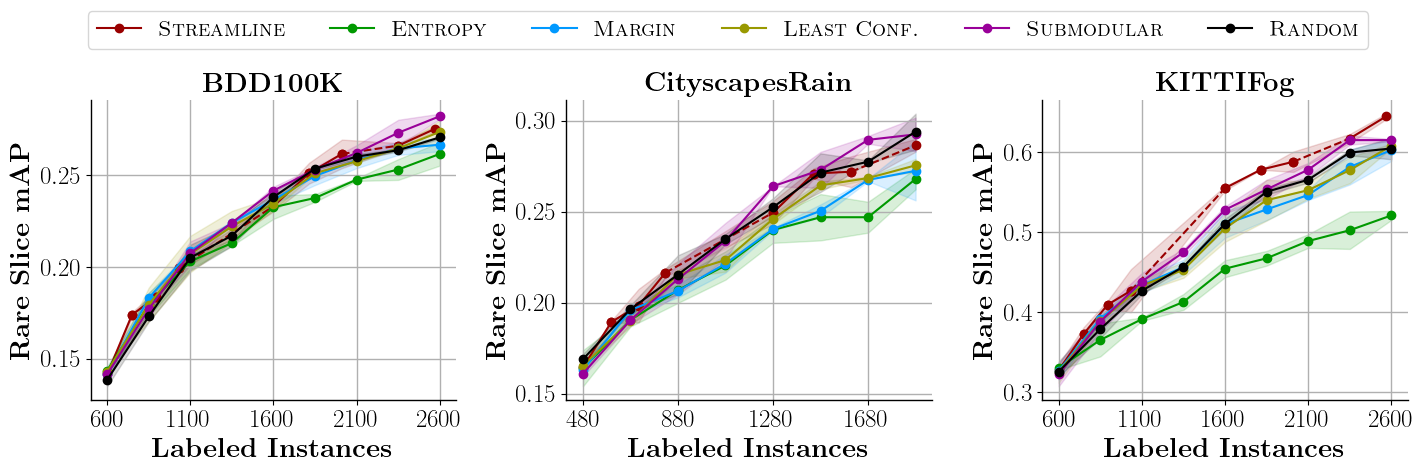}
    \caption{Full mAP of \model\ and the baseline methods from our object detection experiments. \model\ successfully retains comparable full mAP to the baselines while achieving higher rare-slice mAP as shown in~\secref{sec:results}.}
    \label{fig:obj_det_full}
\end{figure}

Here, we present the full accuracy plots achieved by each method studied in~\secref{sec:results} on PovertyMap~\cite{yeh2020using} and iWildCam~\cite{beery2021iwildcam}. Additionally, we present the full mAP plots achieved by each method studied in~\secref{sec:results} on BDD-100K~\cite{yu2020bdd100k}, Cityscapes~\cite{cordts2016cityscapes}, and KITTI~\cite{Geiger2012CVPR}. Each evaluation set is balanced across slices. The full evaluation is given in~\figref{fig:img_cls_full} and in~\figref{fig:obj_det_full}. As shown, \model\ maintains a comparable level of performance to the baselines on the full data while achieving higher rare-slice performance as discussed in~\secref{sec:results}. Notably, \model\ achieves superior performance in settings with particularly hard rare slices of data (such as KITTI's fog slice in~\figref{fig:obj_det_full}).

\section{Size Distribution of Slices}
\label{app:slice_dist}

\begin{figure}[h]
    \centering
    \includegraphics[width=\linewidth]{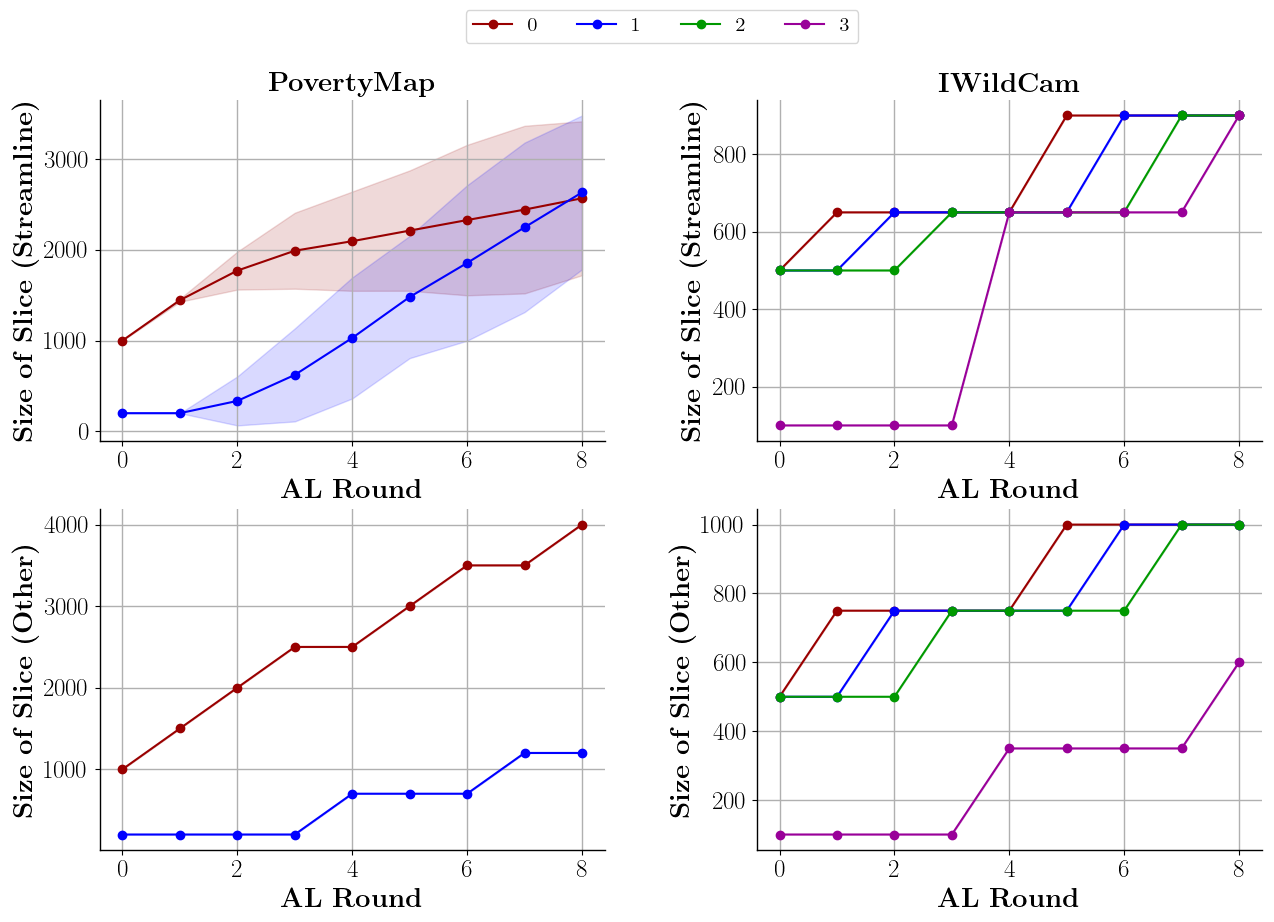}
    \caption{Evolution of slice size in each of our image classification experiments. \model\ correctly identifies the arriving slices in our iWildCam~\cite{beery2021iwildcam} experiment, but it does not always correctly identify the arriving slices in our PovertyMap~\cite{yeh2020using} experiment. However, we note that \model\ does a better job at mitigating slice imbalance than the baselines and that \model\ achieves higher rare-slice accuracy in~\figref{fig:img_cls}. In general, \model\ selects more rare-slice instances than the baselines (other) as shown by a larger "1" slice for PovertyMap and a larger "3" slice for iWildCam.}
    \label{fig:img_cls_dist}
\end{figure}

\begin{figure}[h]
    \centering
    \includegraphics[width=\linewidth]{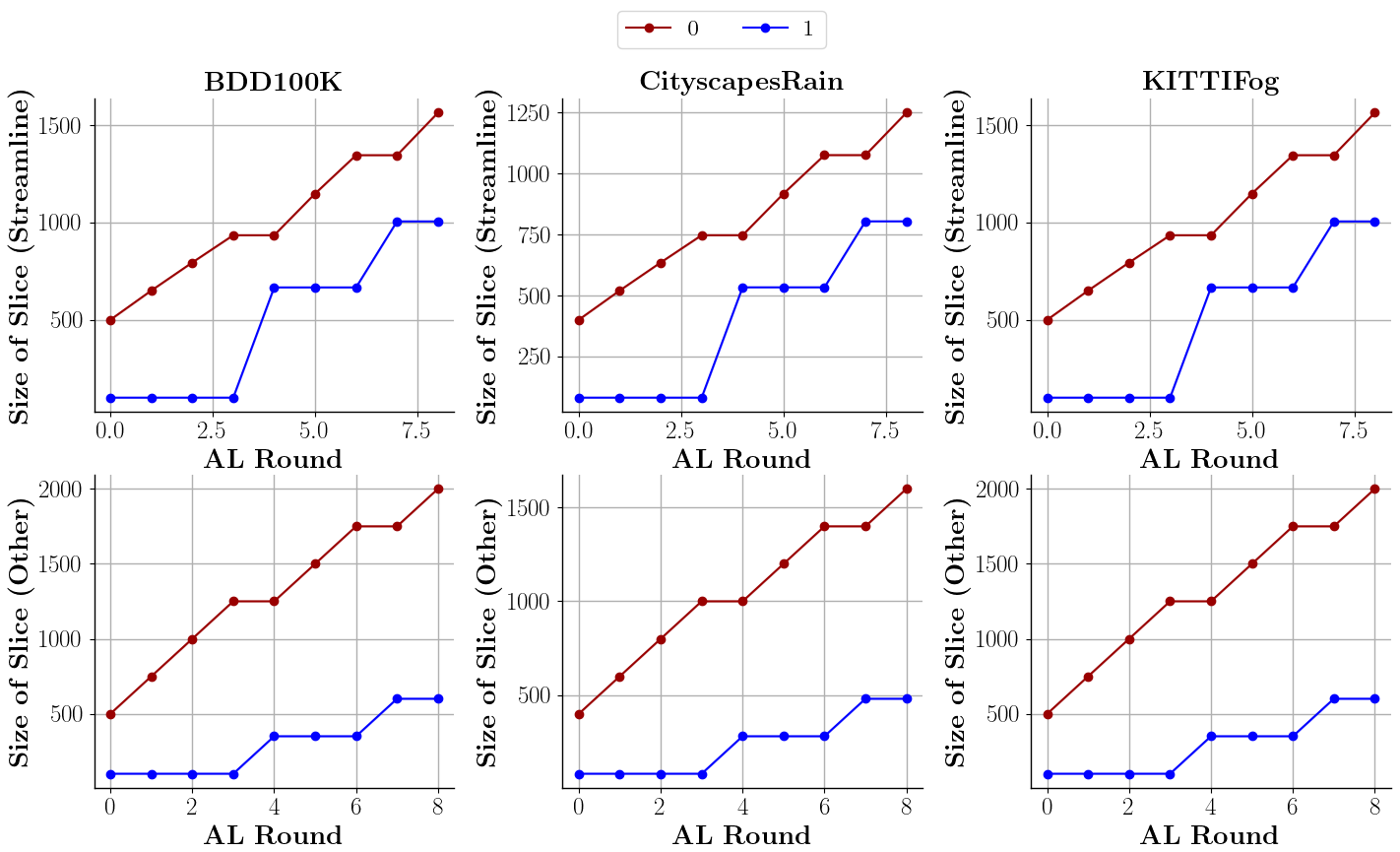}
    \caption{Evolution of slice size in each of our object detection experiments. In all cases, \model\ correctly identifies the arriving slice. Furthermore, \model\ selects more rare-slice instances than the baselines (other) as shown by a larger "1" slice (rare), thereby more effectively mitigating slice imbalance.}
    \label{fig:obj_det_dist}
\end{figure}

Here, we present the size distribution of each slice as it evolves in each of our experiments in~\figref{fig:img_cls_dist} and~\figref{fig:obj_det_dist}. Notably, the rare slice in each experiment is denoted by the largest number in the legend that appears in the plot (\eg, "1" is the rare slice for every dataset except iWildCam~\cite{beery2021iwildcam}, where the rare slice is "3"). Per our experiment design, we introduce each slice sequentially across AL rounds in our iWildCam experiment. In all the other experiments, we introduce the rare slice every 3 rounds. In all but our PovertyMap~\cite{yeh2020using} experiment, \model\ correctly identifies the incoming slice for all selection rounds. In PovertyMap, \model\ does not always identify the incoming slice correctly (as shown by the error bars); however, \model\ mitigates the disparity in size of the rare slice much more effectively than the baselines in every experiment. Hence, \model\ is able to produce models that more effectively handle rare slices of data.

\section{Additional Experiment Details}
\label{app:exp_details}

Here, we list additional details concerning our experiments. 

\subsection{Reproducibility and Licenses}

To reproduce our experiments in \secref{sec:results}, we provide our code respository\footnote{\href{https://github.com/nab170130/new_streamline}{https://github.com/nab170130/new\_streamline}}, which contains detailed instructions for reproducing our results and the exact experiment configurations used in our experiments. In generating our results, we used a local machine with two NVIDIA RTX A6000 GPUs and 96 GB of RAM. Additionally, our code base uses a number of existing assets -- we list their licenses below:

\begin{itemize}
    \item BDD-100K~\cite{yu2020bdd100k}: BSD 3-Clause
    \item Cityscapes~\cite{cordts2016cityscapes}: \href{https://www.cityscapes-dataset.com/license/}{Non-commercial}
    \item DISTIL~\cite{eval}: MIT
    \item iWildCam~\cite{beery2021iwildcam}: MIT (distributed under WILDS)
    \item KITTI~\cite{Geiger2012CVPR}: Creative Commons Attribution-NonCommercial-ShareAlike 3.0
    \item MMDetection~\cite{mmdetection}: Apache 2.0
    \item MMSegmentation~\cite{mmseg2020}: Apache 2.0
    \item PovertyMap~\cite{yeh2020using}: MIT (distributed under WILDS)
    \item PyTorch~\cite{pytorch}: \href{https://github.com/pytorch/pytorch/blob/master/LICENSE}{BSD-style}
    \item Submodlib~\cite{submodlib}: None
    \item WeatherAugment~\cite{halder2019physics}: MIT (code), Creative Commons Attribution-NonCommercial-ShareAlike 3.0 (data)
    \item WILDS~\cite{koh2021wilds}: MIT
\end{itemize}

\subsection{Image Classification Baselines}

For our image classification experiments, we utilize five baseline methods for comparison against \model: \textsc{Badge}~\cite{badge}, \textsc{Entropy}~\cite{alsurvey}, \textsc{Submodular}~\cite{fujii2016budgeted}, SIMILAR~\cite{similar}, and \textsc{Random}. In \textsc{Badge}~\cite{badge}, unlabeled instances are represented by their hypothesized loss gradient embedding at the last fully-connected layer of the model to be trained by using the model's predicted label. Subsequently, \textsc{Badge}~\cite{badge} performs a \textsc{k-means++} initialization upon the space formed by these embeddings to select a diverse set of uncertain instances for labeling as previously detailed in \secref{sec:Intro}. \textsc{Entropy}~\cite{alsurvey} selects the top $k$ instances from the unlabeled data that have the highest entropy over the predicted class probability scores. \textsc{Submodular} selects $k$ instances by \textbf{1)} representing each unlabeled instance by their last-layer features, \textbf{2)} formulating a $|\Ucal| \times |\Ucal|$ cosine similarity kernel $S$ between these representations, \textbf{3)} instantiating the Facility Location (FL) function $F(A) = \sum_{i \in \Ucal} \max_{j \in A} S_{ij}$, and \textbf{4)} using the greedy monotone-submodular maximization algorithm~\cite{nemhauser} on $F$ to formulate the selected set of instances. SIMILAR~\cite{similar} selects $k$ instances by \textbf{1)} generating \textsc{Badge} embeddings for each unlabeled instance and each labeled rare-slice instance in $P_t$, \textbf{2)} formulating a $|\Ucal| \times |P_t|$ cosine similarity kernel $S$ between these representations, \textbf{3)} instantiating the FLQMI function $I_F(A;P_t) = \sum_{i \in P_t} \max_{j\in A} S_{ji} + \sum_{i \in A} \max_{j \in P_t} S_{ij}$, and \textbf{4)} using the greedy monotone-submodular maximization algorithm~\cite{nemhauser} to formulate the selected set of instances. \textsc{Random} simply chooses $k$ instances from the unlabeled data at random.

\subsection{Object Detection Baselines}

For our object detection experiments, we adapt the common uncertainty-based methods of \textsc{Entropy}, \textsc{Margin}, and \textsc{Least Conf.} to the object detection setting by following the scheme of~\cite{kothawade2022talisman}. Specifically, the entropy of each bounding box's class probabilities are average together to yield a final score:

\begin{equation}
    s_{entropy} = \frac{1}{n_{prop}} \sum_{i=1}^{n_{prop}} - \sum_{c=1}^C p^{(i)}_c \log p^{(i)}_c
\end{equation}

\begin{equation}
s_{least\ conf.} = \frac{1}{n_{prop}} \sum_{i=1}^{n_{prop}} 1 - p^{(i)}_{\sigma_1}
\end{equation}

\begin{equation}
s_{margin} = \frac{1}{n_{prop}} \sum_{i=1}^{n_{prop}} p^{(i)}_{\sigma_1} - p^{(i)}_{\sigma_2}
\end{equation}

\noindent where $\sigma_i$ denotes the $i$th highest probability score and $p^{(i)}_c$ denotes the class probability for class $c$ for bounding box prediction $i$. \textsc{Submodular} selects $k$ instances by \textbf{1)} representing each unlabeled instance by their CNN-extracted backbone features, \textbf{2)} formulating a $|\Ucal| \times |\Ucal|$ cosine similarity kernel $S$ between these representations, \textbf{3)} instantiating the Facility Location (FL) function $F(A) = \sum_{i \in \Ucal} \max_{j \in A} S_{ij}$, and \textbf{4)} using the greedy monotone-submodular maximization algorithm~\cite{nemhauser} on $F$ to formulate the selected set of instances.

\section{Limitations, Impacts, and Future Directions for \model}
\label{app:limitations}

Here, we briefly discuss some limitations of \model\ concerning its general use. In our problem setting, we formulate the streaming problem as an \emph{episodic} stream, where slices of the data arrive in a sequence with well-defined boundaries. In many cases, such well-defined boundaries do exist -- such as the autonomous vehicle setting -- but other cases feature \emph{gradual} changes across slices of the data where the unlabeled stream consists of a \emph{mixture} of slices. As \model's selection mechanism is designed for the episodic setting, its selection mechanism may underperform in this non-episodic setting since the unlabeled stream has a mixture of slice identity. As a potential modification, \model\ can be made to sample from the unlabeled stream more frequently to better leverage the temporal locality of data arriving in the stream. Indeed, while the unlabeled stream may be presenting a mixture of slices, it is also likely in many scenarios that data from the same slice are generated in bursts, resulting in temporal locality of the arriving data. Another limitation of \model\ is that its design assumes that each slice category is known \emph{a priori}. In many situations, new distributions of data arise when collecting data from a spatially or temporally evolved source; hence, new slices of the data can arrive in the unlabeled stream. As a possible modification, \model\ can be instrumented with a similarity threshold during its identification step (see \AlgRef{alg:ident}) that, if not broken, can initiate the creation of a new labeled slice of data. Afterwards, the second and third phases of \model\ can be executed as usual to procure data for this new slice. We leave this modification as an avenue of future research. As a concluding remark, we highlight that \model\'s mechanism has the potential to improve the performance of any particular slice(s) by categorizing them as a rare slice. Accordingly, \model\ can improve fairness across slices by targeting those that are underperformant. However, the same mechanism can be used to widen the gap between each slice's performance, which can have negative societal implications if these slices are delineated by cultural aspects such as nationality, age, gender, and so forth. Hence, practitioners should be mindful about slice fairness when applying \model\ in these settings.

\end{document}